\documentclass{article}

\usepackage{PRIMEarxiv}

\usepackage[utf8]{inputenc} 
\usepackage[T1]{fontenc}    
\usepackage{hyperref}       
\usepackage{url}            
\usepackage{booktabs}       
\usepackage{amsfonts}       
\usepackage{nicefrac}       
\usepackage{microtype}      
\usepackage{lipsum}
\usepackage{fancyhdr}       
\usepackage{graphicx}       
\graphicspath{{media/}}     
\usepackage{subcaption}
\usepackage{CJKutf8}

\title{CalliPaint: Chinese Calligraphy Inpainting with Diffusion Model
}

\author{%
  \textbf{Qisheng Liao$^1$,  Zhinuo Wang$^2$, Muhammad Abdul-Mageed$^{1,3}$, Gus Xia$^1$} \\
  $^1$MBZUAI, $^2$Brown University, $^3$The University of British Columbia \\
  \texttt{$^1$\{Qisheng.Liao,Muhammad.Mageed,Gus.Xia\}@mbzuai.ac.ae} \\
  \texttt{$^2$zhinuo\_wang@brown.edu}\\
}

\begin{document}
\maketitle

\begin{abstract}
Chinese calligraphy can be viewed as a unique form of visual art. Recent advancements in computer vision hold significant potential for the future development of generative models in the realm of Chinese calligraphy. Nevertheless, methods of Chinese calligraphy inpainting, which can be effectively used in the art and education fields, remain relatively unexplored. In this paper, we introduce a new model that harnesses recent advancements in both Chinese calligraphy generation and image inpainting. We demonstrate that our proposed model CalliPaint can produce convincing Chinese calligraphy.
\end{abstract}

\begin{figure}[h]
    \centering
    \includegraphics[width=1\textwidth]{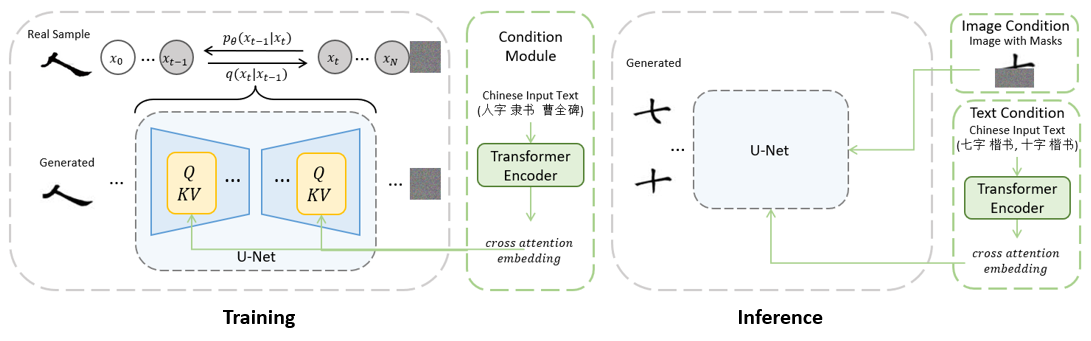}
    \caption{An illustration of our model. The left side displays the structure of our training procedure. During training, every image includes annotated text that showcases the character, script, and style of that specific image. The U-Net models are trained to acquire the ability to produce such images starting from Gaussian noise. The right side illustrates the process of inference. In this instance, we employ two distinct textual conditions while maintaining the same image condition. The resulting outcomes exhibit identical unmasked sections, but the masked portions are generated differently depending on the text conditions.}
    \label{fig:model2}
\end{figure}

\section{Introduction}

Chinese calligraphy is a highly regarded and traditional art form that involves writing Chinese characters with a brush and ink. It is not just a means of communication but a visual art that has deep cultural and historical significance in China and other East Asian countries.
Lately, there has been a noticeable trend in the use of machine learning for the creation of Chinese calligraphy. Examples include  Zi2zi~\cite{zi2zi}, CalliGAN~\cite{wu2020calligan}, and ZiGAN~\cite{wen2021zigan}, which typically employ a Generative Adversarial Network (GAN) architecture. While, Calliffusion~\cite{liao2023calliffusion}, a new proposed work for Chinese calligraphy generation, uses Denoising Diffusion Probabilistic Models(DDPMs)~\cite{ho2020denoising} for generations.

DDPMs generate samples that match the data after a certain amount of time. In the forward diffusion process, a small amount of Gaussian noise is added to a data point sampled from a real data distribution in multiple steps, resulting in a sequence of noisy samples. While in the reverse diffusion process, the images are rebuilt based on the Gaussian distribution. Using the foundation of DDPMs, RePaint~\cite{lugmayr2022repaint} is designed for the task of image inpainting. In RePaint, a novel scheduler is introduced, enabling image inpainting to be performed during the inference stage without the need for additional training.

In this work, we employ the Calliffusion model and RePaint scheduler to design a new framework that can do Chinese calligraphy inpainting with text conditions. This framework could be used not only in the art field but also in education.
\section{Model Architecture}
In training, we follow the settings of Calliffusion. Namely, we use a short description of Chinese text input to control the generations including the characters, scripts, and styles. In the inference phase, we need to provide an extra image with masks to guide the model to fulfill the masked parts. The details of our model architecture are introduced in Figure~\ref{fig:model2}.

\section{Evaluation}
We assess our outcomes using a dual approach, considering both objective and subjective criteria. 

In terms of objective evaluation, we employ the same pre-trained classifier that Calliffusion uses to identify the inpainted images. We choose the same conditions tested in Calliffusion but provide the gold images with random masks as image conditions. The findings presented in Table~\ref{table1} demonstrate that, following the process of inpainting, the image quality from our model maintains a high standard. The classifier also successfully achieves accurate character predictions. Specifically, the average accuracy for inpainted images stands at 0.95, in contrast to the reported accuracy of 0.85 in Calliffusion. This difference in performance can be attributed to the fact that, in Calliffusion, the images under evaluation are entirely generated by the model. Meanwhile, in the case of inpainting, real images are extra conditions and the model only generates the masked parts, which is an easier task.

For subjective evaluation, we design a survey and invite 30 native Chinese speakers to answer. Twelve of them knew Chinese calligraphy before. The results in Table~\ref{tabelsurvey} indicate that it is hard for native Chinese speakers to distinguish the calligraphy inpainted by our model from real calligraphy. The examples of questions in our survey are shown in Figure~\ref{fig:example} in Appendix~\ref{appendix:example}.

\begin{table*}
\centering
\resizebox{\columnwidth}{!}{%
\begin{tabular}{lcccccccccccc}
\hline
                                               & \multicolumn{2}{c}{Regular}                                & \multicolumn{2}{c}{Semi-cursive}                           & \multicolumn{2}{c}{Cursive}                                & \multicolumn{2}{c}{Clerical}                               & \multicolumn{2}{c}{Seal}                                   & \multicolumn{2}{c}{Total}                                  \\
                                               & \multicolumn{1}{l}{Script} & \multicolumn{1}{l}{Character} & \multicolumn{1}{l}{Script} & \multicolumn{1}{l}{Character} & \multicolumn{1}{l}{Script} & \multicolumn{1}{l}{Character} & \multicolumn{1}{l}{Script} & \multicolumn{1}{l}{Character} & \multicolumn{1}{l}{Script} & \multicolumn{1}{l}{Character} & \multicolumn{1}{l}{Script} & \multicolumn{1}{l}{Character} \\ \hline
Real Samples$\ddagger$             & 0.91                       & 0.93                          & 0.83                       & 0.81                          & 0.88                       & 0.68                          & 0.96                       & 0.83                          & 0.97                       & 0.81                          & 0.88                       & 0.78                          \\

Calliffusion$\ddagger$  & 0.96                       &  0.94 & 0.86                       & 0.95                          & 0.88                       & 0.64                          & 0.97                       & 0.91                 & 0.99              & 0.79                          & 0.93                       & 0.85   \\  
CalliPaint & 0.98                       & 0.96                          & 0.95                       & 0.99                          & 0.99 & 0.93                          & 0.99                       & 0.98                          & 0.99                       & 0.90                 & 0.98                       & 0.95   \\\hline                        
\end{tabular}%
}
\caption{The performance of our generated data in different scripts in accuracy. $\ddagger$ indicate the results from Calliffusion.}
\label{table1}
\end{table*}

\begin{table}[]
\centering
\begin{tabular}{lcccccc}
\hline
       & \multicolumn{2}{c}{People know Calligraphy} & \multicolumn{2}{c}{People do not know Calligraphy} & \multicolumn{2}{c}{Total} \\
       & Acc                 & P-Value               & Acc                     & P-Value                    & Acc        & P-Value      \\ \hline
Type 1 & 0.28                & 0.94                  & 0.22                    & 0.94                       & 0.24       & 0.99         \\
Type 2 & 0.04                & 0.63                  & 0.08                    & 0.69                       & 0.06       & 0.66         \\ \hline
\end{tabular}
\caption{The accuracy and p-value of each type of question in our survey.}
\label{tabelsurvey}
\end{table}

\section{Conclusion}
This paper represents a stride in the direction of employing diffusion models for Chinese calligraphy inpainting. The utility of inpainting extends across diverse domains. Nevertheless, we are currently confronted with certain drawbacks, notably the relatively slow inference time. We anticipate the resolution of this issue in forthcoming developments. 

\section*{Ethical Implications}
The ethical considerations of this project involve the intention to offer users an efficient tool for restoring missing elements in calligraphy images, which has broad applications such as correcting the inappropriate part of new learners' calligraphy and making it become an artwork written by famous calligraphers. The training data exclusively originate from publicly available websites, and the calligraphy art itself is written by ancient Chinese calligraphers.

\bibliographystyle{unsrt}  
\bibliography{references}  
\appendix
\section{Qualitative Results}
\begin{figure}[h]
    \centering
    \includegraphics[width=1\textwidth]{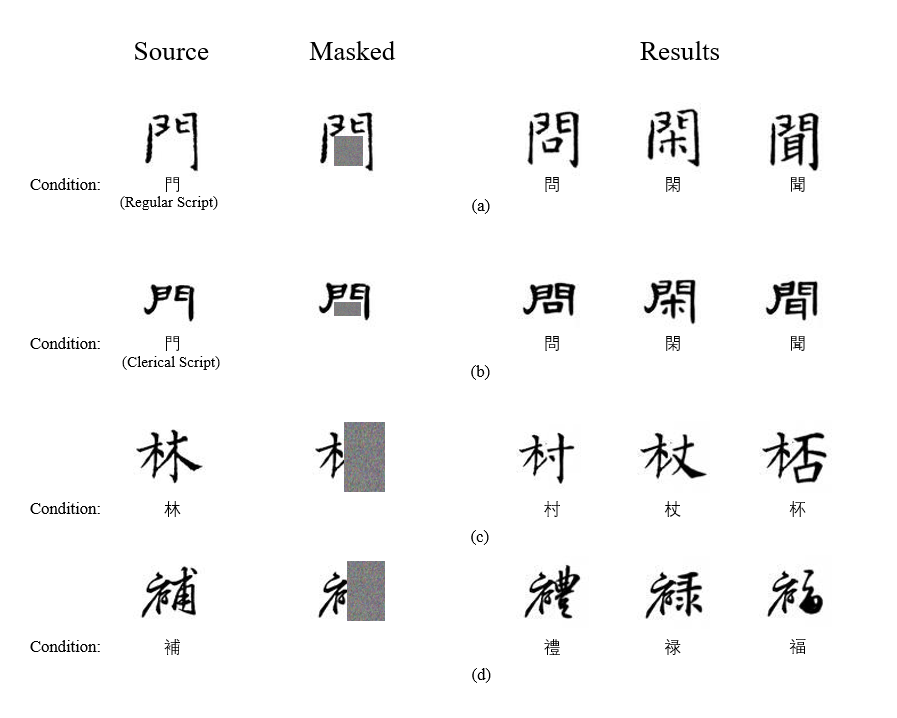}
    \caption{The qualitative results of our model.}
    \label{fig:qualitative}
\end{figure}

The qualitative results are shown in Figure \ref{fig:qualitative}. In Figure \ref{fig:qualitative}(a) and Figure \ref{fig:qualitative}(b), we show the generation of the same character in different scripts.In \ref{fig:qualitative}(c), we show the generation with some flaws. In \ref{fig:qualitative}(d), we show the generation of non-existing characters.

\begin{figure}[!htb]
     \centering
     \begin{subfigure}[b]{0.45\textwidth}
         \centering
         \includegraphics[width=\textwidth]{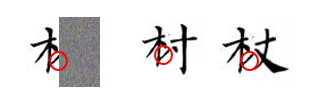}
         \caption{The generation with flaws. The red circles denote the flaw, an extra stroke that comes from the unmasked part of the original image.}
         \label{fig:miss}
     \end{subfigure}
     \hfill
     \begin{subfigure}[b]{0.45\textwidth}
         \centering
         \includegraphics[width=\textwidth]{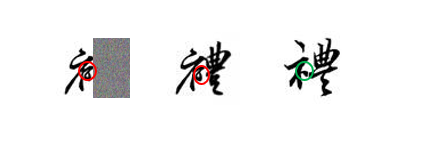}
         \caption{The generation of non-existing characters. The red circles denote the flaw. The image with the green circle is the correct existing character. }
         \label{fig:add}
     \end{subfigure} 
     \caption{The generation with flaws and non-existing characters.}
    \label{fig:flaw}
\end{figure}

The flaw of Figure \ref{fig:qualitative}(c) is shown in Figure \ref{fig:flaw}(a). There is an extra stroke left in the generation that comes from the unmasked part of the original image. Based on this observation, we intentionally manipulate the generation to force the model to generate some non-existing character shown in Figure \ref{fig:qualitative}(d). In Figure \ref{fig:flaw}(b), we use red circles and a green circle to compare the non-existing character and the correct one. \begin{CJK*}{UTF8}{gbsn}More specifically, the left part of the character 禮 in our example shoule be "礻" but we intentionally use "衤" as the image condition to inpainting.\end{CJK*} Based on this setting, the model generate the image that does not exist.

\section{Examples of Subjective Surveys}
\label{appendix:example}
\begin{figure}[!htb]
     \centering
     \begin{subfigure}[b]{0.45\textwidth}
         \centering
         \includegraphics[width=\textwidth]{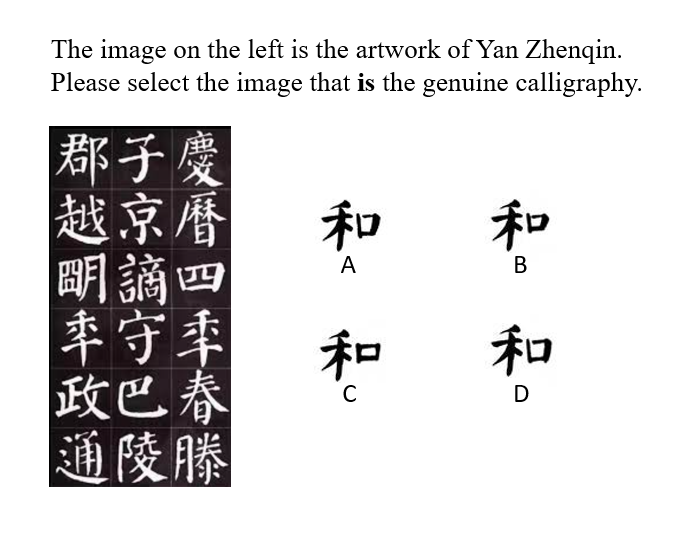}
         \caption{Question that asks the respondent to find the genuine calligraphy.}
         \label{fig:example1}
     \end{subfigure}
     \hfill
     \begin{subfigure}[b]{0.45\textwidth}
         \centering
         \includegraphics[width=\textwidth]{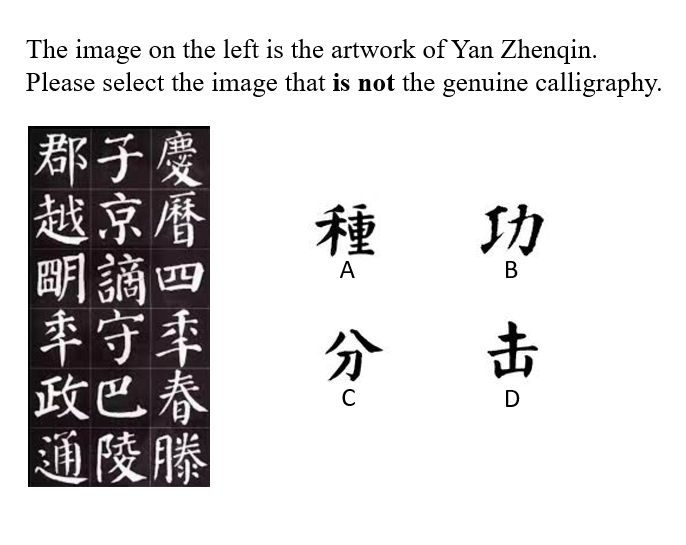}
         \caption{Question that asks the respondent to find the fake calligraphy.}
         \label{fig:example2}
     \end{subfigure} 
     \caption{Two example questions in our surveys.}
    \label{fig:example}
\end{figure}

In Figure \ref{fig:example}, we show two example questions in our survey. Figure \ref{fig:example}(a) is the question that asks respondents to find genuine calligraphy and Figure \ref{fig:example}(b) asks respondents to find the fake one. The correct answer for both questions is Option A.

\end{document}